\documentclass[10pt,twocolumn,letterpaper]{article}

\usepackage{iccv}
\usepackage{times}
\usepackage{epsfig}
\usepackage{graphicx}
\usepackage{amsmath}
\usepackage{amssymb}
\usepackage{algorithm}
\usepackage{algorithmic}
\usepackage[numbers,sort]{natbib}

\usepackage[pagebackref=true,breaklinks=true,letterpaper=true,colorlinks,bookmarks=false]{hyperref}

\iccvfinalcopy 
\def\iccvPaperID{3102} 

\ificcvfinal\fi
\begin{document}

\title{Learning Human-Object Interaction Detection using Interaction Points}

\newcommand{\sep}{\quad}

\author{Tiancai Wang$^{1}$\thanks{This work was performed during an internship at MEGVII Technology. This work was supported by The National Key Research and Development Program of China (No. 2017YFA0700800) and Beijing Academy of Artificial Intelligence (BAAI). Contact email: wangtc@tju.edu.cn }  \qquad Tong Yang$^{1}$ \qquad Martin Danelljan$^{2}$ \qquad Fahad Shahbaz Khan$^{3,4}$\\ Xiangyu Zhang$^{1}$ \qquad Jian Sun$^1$\vspace{1mm}\\
$^1$MEGVII Technology \sep
$^2$ETH Zurich, Switzerland \sep
$^3$IIAI, UAE \sep
$^4$Link\"oping University, Sweden
}

\maketitle
\ificcvfinal\thispagestyle{empty}\fi


\begin{abstract}
Understanding interactions between humans and objects is one of the fundamental problems in visual classification and an essential step towards detailed scene understanding. Human-object interaction (HOI) detection strives to localize both the human and an object as well as the identification of complex interactions between them.  Most existing HOI detection approaches are instance-centric where interactions between all possible human-object pairs are predicted based on appearance features and coarse spatial information. We argue that appearance features alone are insufficient to capture complex human-object interactions. In this paper, we therefore propose a novel fully-convolutional approach that directly detects the interactions between human-object pairs. Our network predicts \emph{interaction points}, which directly localize and classify the interaction. Paired with the densely predicted \emph{interaction vectors}, the interactions are associated with human and object detections to obtain final predictions. To the best of our knowledge, we are the first to propose an approach where  HOI detection is posed as a keypoint detection and grouping problem. Experiments are performed on two popular benchmarks: V-COCO and HICO-DET. Our approach sets a new state-of-the-art on both datasets. Code is available at \url{https://github.com/vaesl/IP-Net}. 

\vspace{-0.3cm}
\end{abstract}

\section{Introduction}
Detailed semantic understanding of image contents, beyond instance-level recognition, is one of the fundamental problems in computer vision. Detecting human-object interaction (HOI) is a class of visual relationship detection where the task is to not only localize both a human and an object but also infer the relationship between them, such as ``eating an apple'' or ``driving a car''. The problem is challenging since an image may contain multiple humans performing the same interaction, same human simultaneously interacting with multiple objects (``sit on a couch and type on laptop''), multiple humans sharing the same interaction and object (``throw and catch ball''), or fine-grained interactions (``walk horse'', ``feed horse'' and ``jump horse''). These complex and diverse interaction scenarios impose significant challenges when designing an HOI detection solution.


\begin{figure}[t]
\begin{center}
   \includegraphics[width=\linewidth]{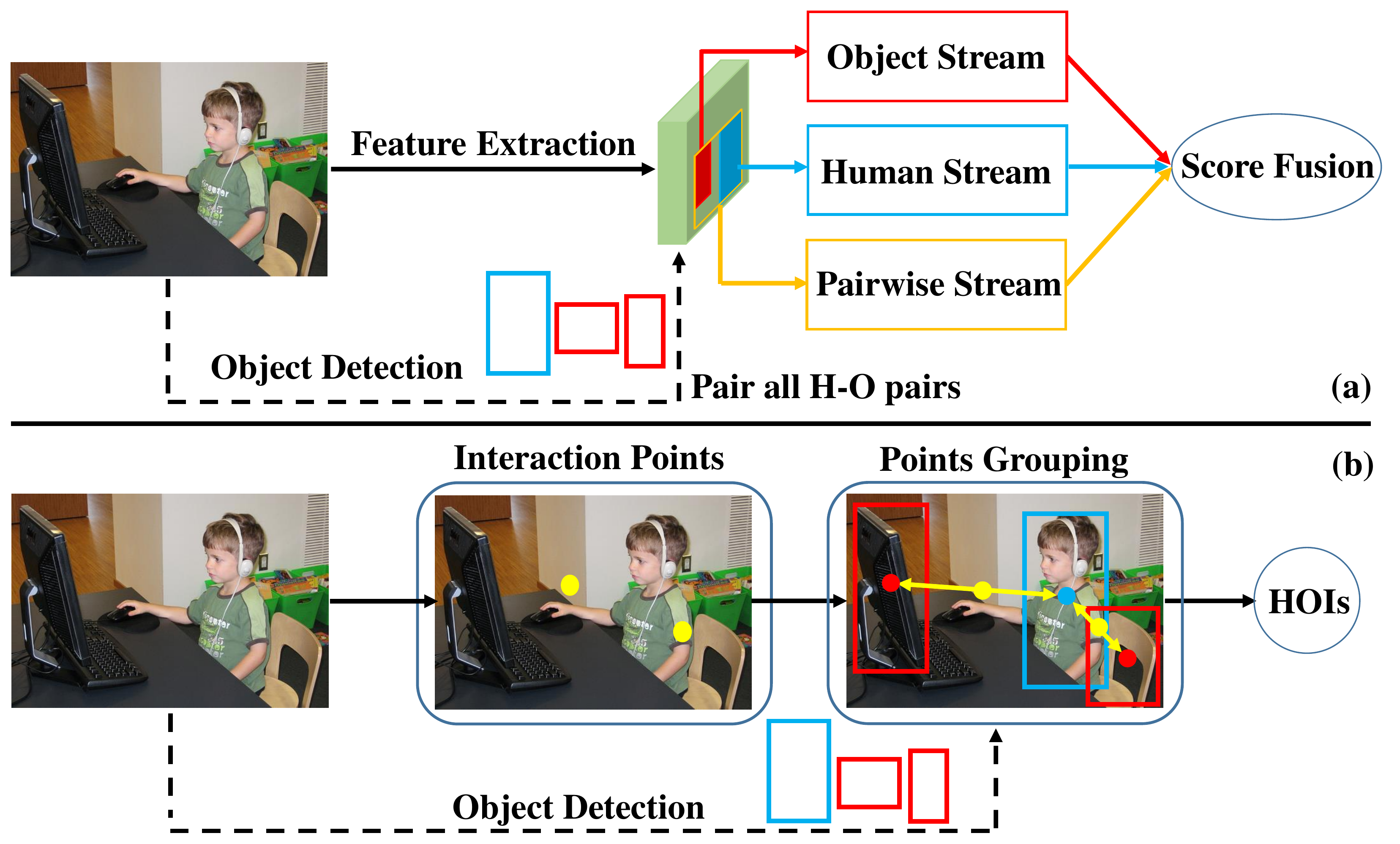}
\end{center} \vspace{-0.4cm}
   \caption{(a) Most existing approaches address the HOI detection problem where the detected bounding-boxes (human and object) from a pre-trained detector are first used to extract region-of-interest (RoI) features from the backbone. Then, a multi-stream architecture is employed where the individual scores from three parallel streams: human, object and pairwise are fused to obtain final interaction predictions for all human-object pairs. (b) Different from previous methods, our proposed approach poses HOI as a keypoint detection and grouping problem by learning to generate interaction points and vectors which are directly grouped along with human and object instances from the object detection branch.
   } \vspace{-0.3cm}
\label{method_compare_intro}
\end{figure}


Most existing approaches \cite{gao2018ican, li2019TIN, chao2018hoi, zhou2020} detect human-object interactions in the form of triplets $\langle$\textit{human, action, object}$\rangle$ by decomposing the problem into two parts: object detection and interaction recognition. For object detection, a pre-trained object detector is typically employed to detect both humans and objects. For interaction recognition, several strategies exist in literature \cite{qi2018learning, wang2019iccv, Peyre_ICCV19}. Most of the recent HOI detection approaches \cite{chao2018hoi, gao2018ican, li2019TIN, wang2019iccv} utilize a multi-stream architecture (see Fig.~\ref{method_compare_intro}(a)) for interaction recognition. The multi-stream architecture typically contains three individual streams: a human, an object, and a pairwise. Both human and object streams encode appearance features of human and objects, respectively whereas the pairwise stream aims at encoding the spatial relationship between the human and object.  Individual scores from the three streams are then fused in a late fusion fashion for interaction recognition.

While improving the HOI detection performance, state-of-the-art approaches based on the above mentioned multi-stream architecture are computationally expensive. During training, these instance-centric approaches require pairing all humans with all objects in order to  learn both positive and negative human-object pairs. This implies that the inference time scales \emph{quadratically} with the number of instances in the scene, since all human-object pairs are required to be passed through the network in order to obtain the final interaction scores. In addition to being computationally expensive, these approaches predominantly rely on appearance features and a simple pairwise stream that takes the union of the two boxes (human and object) to construct a binary image  representation. We argue that this reliance on appearance features alone and coarse spatial information is insufficient to capture complex interactions, leading to inaccurate predictions. In this work, we look into an alternative approach that addresses these shortcomings by directly detecting the interactions  between human-object pairs as a set of interaction points. 

\noindent \textbf{Contributions:} In this work, we propose a novel approach for HOI detection. Motivated by the recent success of anchor-free object detection methods, we pose HOI detection as a keypoint detection and grouping problem (see Fig.~\ref{method_compare_intro}(b)). The proposed approach directly detects interactions between human-object pairs as a set of interaction points. Based on the interaction point, our method learns to generate an interaction vector with respect to the human and object center points. We further introduce an interaction grouping scheme that pairs the interaction point and vector with the corresponding human and object bounding-box predictions, from the detection branch, to produce final interaction predictions.
Extensive experiments are conducted on two HOI detection benchmarks: V-COCO \cite{gupta2015vcoco} and HICO-DET \cite{chao2015hico} datasets. Our proposed architecture achieves state-of-the-art results on both two datasets, outperforming existing instance-centric methods by a significant margin. Additionally, we perform a thorough ablation study to demonstrate the effectiveness of our approach.

\section{Related Work}

\noindent \textbf{Object Detection:} In recent years, significant progress has been made in the field of object detection \cite{fasterrcnn2015, lin2017fpn, WLiu2016SSD, KhanTIP15, wang2019learning,wang2018attentive, MGAN_2019_ICCV, nie2019efgr}, mainly due to the advances in deep convolutional neural networks (CNNs). Generally, modern object detection approaches can be divided into single-stage \cite{WLiu2016SSD,YOLOv2,YOLOv1,EFIP2019,wang2019lrfnet} and two-stage methods \cite{fasterrcnn2015, lin2017fpn, TripleNet2018}. Two-stage object detection methods typically generate candidate object proposals and then perform classification and regression of these proposals in the second stage. On the other hand, single-stage object detection approaches work by directly classifying and regressing the default anchor box in each position. Two-stage object detectors are generally known to be more accurate whereas the main advantage of single-stage methods is their speed.

Within object detection, recent anchor-free single-stage detectors \cite{Cornernet,ExtremeNet,Zhou&CenterNet, FCOS} 
 aim at eliminating the requirement of anchor boxes and treat object detection as keypoint estimation. CornerNet \cite{Cornernet} detects the bounding-box of an object as a pair of keypoints, the top-left corner and the bottom-right corner. ExtremeNet \cite{ExtremeNet} further detects four extreme points and one center point of objects and groups the five keypoints into a bounding-box. CenterNet \cite{Zhou&CenterNet} models an object as a single point — the center point of its bounding-box and is also extended to Human pose estimation \cite{fang2017rmpe} and 3D detection task \cite{Charles2018PointNets}.


\noindent \textbf{Human-Object Interaction Detection} Among existing human-object interaction (HOI) detection methods, the work of \cite{gupta2015vcoco} is the first to explore the problem of visual semantic role labeling. The objective of this problem is to localize the agent (human) and object along with detecting the interaction between them. The work of \cite{gkioxari2017interactnet} introduces a human-centric approach, called InteractNet, which extends the Faster R-CNN framework with an additional branch to learn the interaction-specific density map over target locations. Qi \etal, \cite{qi2018learning} proposes to utilize graph convolution neural network and regards the HOI task as a graph structure optimization problem. Chao \etal, \cite{chao2018hoi} builds a multi-stream network that is based on the human-object region-of-interest and the pairwise interaction branch. The inputs to this multi-stream architecture are the predicted bounding-boxes from the pre-trained detector (\eg, FPN \cite{lin2017fpn}) and the original image. Human and object streams in such a multi-stream architecture are based on appearance features, extracted from the backbone network, to generate confidence predictions on the detected human and object bounding-boxes. The pairwise stream, on the other hand, simply encodes the spatial relationship between the human and object by taking  the  union  of  the  two  boxes (human and object). Later works have extended the above mentioned multi-stream architecture by, \eg, introducing instance-centric attention \cite{gao2018ican}, pose information \cite{li2019TIN} and deep contextual attention based on context-aware appearance features \cite{wang2019iccv}.



\begin{figure*}[t]
\begin{center}
   \includegraphics[width=\linewidth]{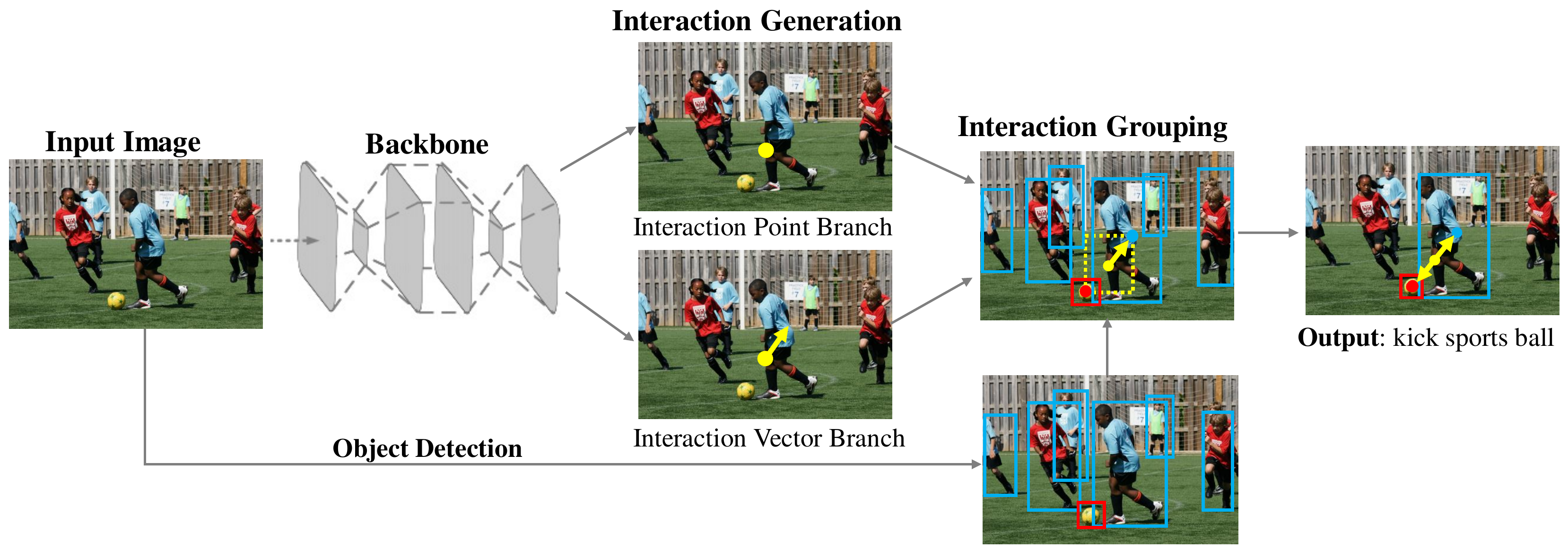}
\end{center} \vspace{-0.4cm}
  \caption{Overall architecture of the proposed HOI detection framework having a localization and an interaction prediction stage. As in several previous works \cite{gao2018ican,wang2019iccv,li2019TIN}, we adopt a standard object detector (FPN~\cite{lin2017fpn}) to obtain human and object bounding-box predictions. Our interaction prediction stage consists of three steps: feature extraction, interaction generation (Sec.~\ref{Interaction_Points_Generation}) and interaction grouping (Sec.~\ref{Interaction_Points_Grouping}). The interaction generation contains two independent branches to produce interaction point and interaction vector, respectively. Interaction point and vector together with the detected human and object bounding-box predictions are then input to the interaction grouping for final HOI predictions: $\langle$\textit{human, action, object}$\rangle$. 
  }  \vspace{-0.2cm}
\label{architecture}
\end{figure*}

\section{Method}
Here, we present our approach based on interaction point generation (Sec.~\ref{Interaction_Points_Generation}) and grouping (Sec.~\ref{Interaction_Points_Grouping}).    



\subsection{Motivation}
As discussed earlier, most existing HOI detection approaches~\cite{gao2018ican,chao2018hoi,wang2019iccv} adopt a multi-stream architecture where individual scores from a human, an object, and a pairwise stream are fused in a late fusion manner for interaction recognition. We argue that such a late fusion strategy is sub-optimal since appearance features alone are insufficient to capture complex human-object interactions. Further, the pairwise stream  simply takes the union of the two boxes (human and object) as the reference box to construct a binary image representation which may lead to inaccurate predictions due to the coarse spatial information. 


Motivated by the advances in anchor-free object detection~\cite{Zhou&CenterNet, Cornernet, ExtremeNet}, we regard HOI detection as interaction point estimation problem by defining the interaction between the human and an object as an interaction point. Based on the interaction point, our method also learns to produce an interaction vector with respect to the human and object center points. It then pairs the interaction points with the corresponding human and object bounding-box predictions. Different from object detection where object instances are generally independent to each other in an image, interaction point estimation in HOI is more challenging due to diverse and complex real-world interaction scenarios, \eg, multiple humans performing the same interaction or same human simultaneously interacting with multiple objects. To the best of our knowledge, we are the first to propose a HOI detection approach where interaction between the human and an object is defined as a keypoint.

\subsection{Overall Architecture}
Our overall architecture is shown in Fig.~\ref{architecture}. It consists of object detection and interaction prediction. For object detection, we follow previous HOI detection works~\cite{gao2018ican,wang2019iccv} and employ a standard
object detector, FPN~\cite{lin2017fpn}, for generating bounding-boxes for all possible
human and object instances in an image. The main focus of our design is a new representation for \textit{interaction prediction}. It comprises three steps: feature extraction, interaction generation (Sec.~\ref{Interaction_Points_Generation}) and interaction grouping (Sec.~\ref{Interaction_Points_Grouping}). For feature extraction, we employ the Hourglass~\cite{hourglass} as the network backbone typically used in anchor-free single stage methods \cite{Cornernet,Zhou&CenterNet,ExtremeNet}. Given an input RGB image with size $H\times W \times 3$, the output of the Hourglass network is a feature map with size $\frac{H}{S}\times \frac{W}{S} \times D$, where $H$, $W$ are the height, width of the input image and $D$, $S$ are the output channels and stride, respectively. As in~\cite{Realtimepose, accuratepose}, we adopt a stride of $S=4$ to achieve a trade-off between accurate localization and computational efficiency. The resulting features from the backbone are input to the interaction generation module to produce interaction point and interaction vector. Interaction point is defined as the center point of the action between a human-object pair and is the starting point of the interaction vector. Consequently, the interaction point and vector together with the detected human and object bounding-boxes are input to the interaction grouping step for the final HOI triplet $\langle$\textit{human, action, object}$\rangle$ prediction. 
\subsection{Interaction Generation}
\label{Interaction_Points_Generation}
The interaction generation module contains two parallel branches: interaction point and interaction vector prediction. Both branches take the features extracted from the backbone as an input.


\begin{figure}[t]
\begin{center}
   \resizebox{0.40\textwidth}{!}{\includegraphics{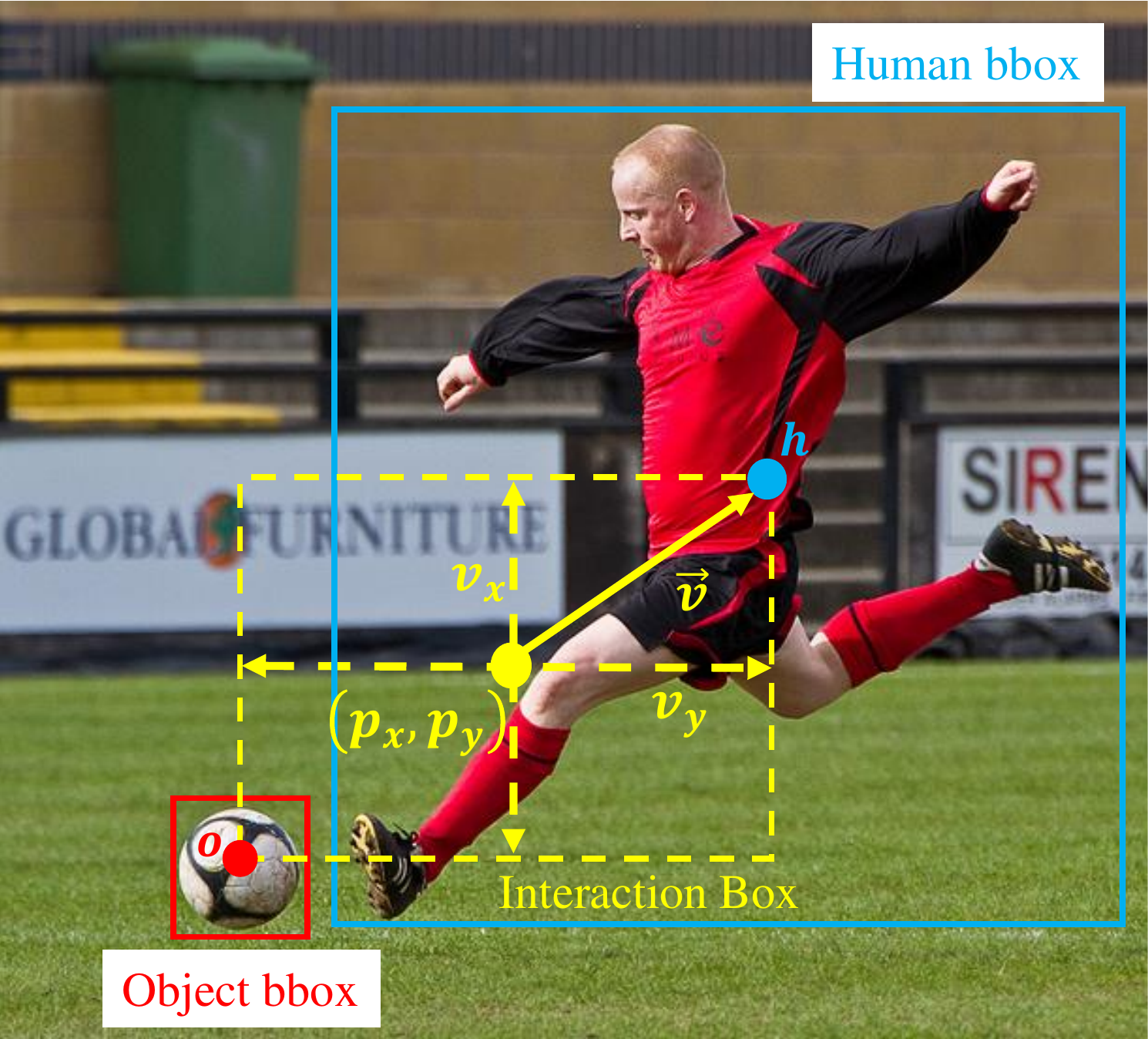}}%
\end{center} \vspace{-0.3cm}
   \caption{Illustration of interaction point and interaction vector on an example image. The interaction  point $p$ (yellow circle) is defined  as the  center point of the action between human-object pair and itself serves as the starting point of the interaction vector $v$ (yellow arrow). During training, interaction point heatmaps are supervised by ground-truth Gaussian heatmaps generated from human and object center points (cyan and red circles). The raw-pixel coordinates of interaction points on the scale maps are supervised by horizontal and vertical lengths
 of the interaction vector using L1 loss. At inference, the generated interaction point heatmaps are used to extract top-k peak interaction points by employing a post-processing strategy, as in \cite{Cornernet}. Based on the location of top-k interaction points, horizontal and vertical length of the interaction vector is obtained at corresponding coordinates of the scale maps.
   } \vspace{-0.2cm}
\label{center_point}
\end{figure}

\noindent \textbf{Interaction Point Branch:} Given the feature maps generated from the backbone network, a single $3\times 3$ convolution layer is employed to produce the interaction point heatmaps of size $\frac{H}{S}\times \frac{W}{S} \times C$, where $C$ denotes the number of interaction categories. During training, the interaction point heatmaps are supervised by the ground-truth heatmaps with multiple peaks, where each interaction point is defined with the same Gaussian Kernel, as in~\cite{Cornernet}. We empirically fix the standard deviation in the Gaussian Kernel throughout our experiments. Note that a single keypoint location can only represent one object class in single-stage object detection~\cite{Zhou&CenterNet}. Different from object detection, a single keypoint location may refer to multiple interaction categories in HOI detection since the human can have multiple interactions with a given object. For instance, the human may hold and hit with tennis racket at the same time. In such a case, both the 'hit' and 'hold' interactions are located at the same position on the heatmap but are represented by different channels. Fig.~\ref{center_point} shows an example interaction point (yellow circle), defined as $p_{x}=\frac{h_{x}+o_{x}}{2},p_{y}=\frac{h_{y}+o_{y}}{2}$, for a given human-object (HO) pair having center points $h=(h_{x},h_{y})$ and $o=(o_{x},o_{y})$, respectively. Note that interaction points are generated for categories involving both the human and an object. For interaction categories without any associated object (\eg, walk and run), the interaction point generalizes to the center point of the corresponding human. Most categories in standard HOI detection datasets~\cite{gupta2015vcoco, chao2015hico} involve both the human and an object. 




\noindent \textbf{Interaction Vector Branch:} 
As shown in Fig.~\ref{center_point}, based on the interaction point $(p_{x},p_{y})$, the interaction vector branch aims to predict the interaction vector towards the corresponding human center point. Given the paired human and object bounding-boxes, human center point $h$, and object center point $o$, the interaction point $p=(p_{x},p_{y})$ is calculated. Then, the interaction vector $v=(v_{x},v_{y})$ is defined such that $p+v=h$ and $p-v=o$. 

The interaction vector branch is trained to predict the value of the unsigned interaction vector $v'=(|v_{x}|,|v_{y}|)$, which is used as the ground-truth in our training. As in the interaction point branch, we employ a single $3\times 3$ convolution layer to produce the unsigned interaction vector map $V$ of size $\frac{H}{S}\times \frac{W}{S} \times 2$, where one is for the length of interaction vector in horizontal direction and the other is for the length of interaction vector in vertical direction.
At inference, we extract four possible locations of the human center based on the interaction point and the unsigned interaction vector as,
\begin{equation}
\label{eq:corners}
    (x_h^i, y_h^i) = (p_x \pm |v_{x}|, p_y \pm |v_{y}|)\,,\; i = 1,2,3,4 \,.
\end{equation}
We further define the \emph{interaction box} as the rectangle with corners given by \eqref{eq:corners}. Next, we describe the interaction grouping scheme.

\subsection{Interaction Grouping}
\label{Interaction_Points_Grouping}
During training, the interaction point and its corresponding human and object center points have a fixed geometric structure. During the inference stage, the generated interaction points need to be grouped with the object detection results (human and object bounding-boxes). This implies that the generated interaction point $p$ is paired with the human having center $h$ and object having center $o$, if the following condition is satisfied: $h \approx p + v$ and $o \approx p - v$.


\begin{figure}[h]
\begin{center}
   \resizebox{0.40\textwidth}{!}{\includegraphics{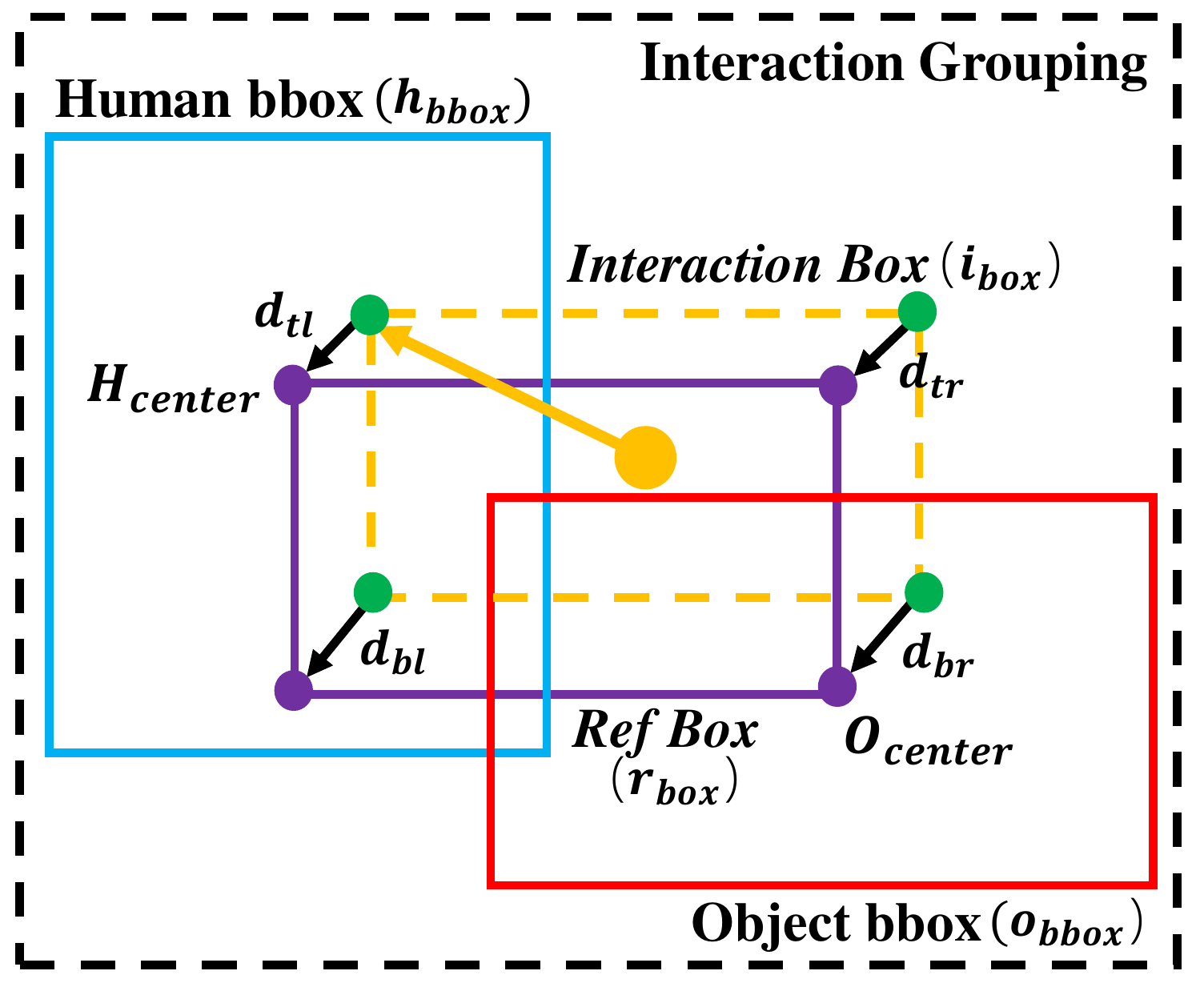}}%
\end{center} \vspace{-0.3cm}
   \caption{The procedure of interaction grouping scheme. It has three inputs: the human/object bounding-boxes from object detection branch, the interaction points from the interaction point branch and the interaction vector predicted by the interaction vector branch. The interaction box (orange box) is determined by the given interaction point and the lengths (horizontal and vertical) of the interaction vector. The reference box (purple) can also determined by the detected human/object boxes. In case the reference box, the interaction box and human/object bounding-boxes satisfy the conditions~\eqref{eq:conditions}, then the current human/object bounding-boxes and the interaction point are regarded as a true positive HO pair.
   } \vspace{-0.3cm}
\label{points_grouping}
\end{figure}

For efficient and accurate grouping of interaction points with human and object bounding-boxes, we further propose an interaction grouping scheme which utilizes soft constraints to filter out bulk of the negative HOI pairs. 
Fig.~\ref{points_grouping} shows an illustration of our interaction grouping scheme. It has three inputs: human/object bounding-box (cyan and red), interaction point (orange point) extracted from interaction heatmaps, and the interaction vector (orange arrow) at the location of interaction point. The four corners (in green) of the interaction box (in orange) are calculated by the given interaction point and the unsigned interaction vector, using Eq.~\eqref{eq:corners}. The four corners of the \emph{reference box} $r_{box}$ (in purple) can be determined by the center points of the detected human and object bounding-boxes. Then, based on the generated interaction and reference boxes, we compute the vector lengths, $d_\text{tl},d_\text{tr},d_\text{bl},d_\text{br}$, for four corners of these two boxes. In case the interaction box and the four vector lengths  satisfy the constraints in~\eqref{eq:conditions} below, then the current human and object bounding-boxes and the interaction point are regarded as the  true positive HOI pair.
%
%
%

\begin{equation}
\label{eq:conditions}
\begin{cases}
     \text{IoU}(h_\text{bbox}, i_\text{box})>0,\\[1ex]
    \text{IoU}(o_\text{bbox}, i_\text{box})>0, \\[1ex]
    d_\text{tl},d_\text{tr},d_\text{bl},d_\text{br}<d_{\tau}
\end{cases}
\end{equation}
Here, $h_\text{bbox}$ and $o_\text{bbox}$ are the given human and object bounding-boxes from object detection branch. $i_{box}$ is the interaction box generated by the interaction point and the interaction vector. $d_\text{tl}$, $d_\text{br}$, $d_\text{tr}$ and $d_\text{bl}$ are four vector lengths of the four corners between interaction box $i_\text{box}$ and the reference box $r_\text{box}$. $d_{\tau}$ is the vector length threshold set for filtering the negative HOI pairs. Interaction grouping scheme is presented in Algorithm~\ref{Interaction_Point_Grouping}.


\begin{algorithm}
\caption{Interaction Grouping} 
\label{Interaction_Point_Grouping}
\begin{algorithmic}

\REQUIRE ~~\\ 
Human/object bboxes from object detector: $H_\text{bbox}$, $O_\text{bbox}$\\Interaction point and vector heatmaps: $P$, $V$\\Human, object and action score thresholds:$h_{\tau}, o_{\tau}, a_{\tau}$\\Vector length threshold for corners: $d_{\tau}$
\ENSURE 
HOI triplets with final score.\\
// Convert heatmaps $P$ into an interaction point set $A$.\\
// Extract the interaction vectors from $V$.
\FOR{$h_\text{box} \in H_\text{bbox}$, $o_\text{box} \in O_\text{bbox}$, $a \in A$}

\IF{$h_\text{score} > h_{\tau}, o_\text{score} > o_{\tau}, a_\text{score} > a_{\tau}$}

\STATE // Obtain interaction box $i_\text{box}$ using Ed.~\ref{eq:corners}.\\
\STATE // Calculate reference box $r_\text{box}$ by $h_\text{box}$ and $o_\text{box}$.\\

\IF{$h_\text{bbox}$, $o_\text{bbox}$, $i_\text{box}$, $r_\text{box}$ satisfy Condition~\eqref{eq:conditions}}

\STATE  $s_\text{f} \leftarrow h_\text{score}\cdot o_\text{score} \cdot p_\text{score}$
\STATE // Output the current HOI pair with final score $s_\text{f}$ .\\

\ENDIF
\ENDIF
\ENDFOR
\end{algorithmic}
\end{algorithm}
\subsection{Model Learning}
For the predicted interaction point heatmap $P$, the ground-truth heatmaps $\hat{P}$ are with all interaction points, and each of them is defined as the Gaussian Kernel. We follow the modified focal loss originally proposed in \cite{Cornernet} to balance the positive and negative samples,
\begin{equation}
\label{loss_for_point}
L_{p}=\frac{-1}{N_{p}}\left\{
\begin{aligned}
(1-P_{xyc})^\alpha\log(P_{xyc}), \hspace{1.8em} \text{if}~\hat{P}_{xyc}=1 \\
(1-\hat{P}_{xyc})^\beta(P_{xyc})^\alpha\log(1-P_{xyc}), \text{o.w.}
\end{aligned}
\right.
\end{equation}
%
where $N_{p}$ is the number of interaction points in the image. $\alpha$ and $\beta$ are the hyper-parameters to control the contribution of each point (we set $\alpha$ to 2 and $\beta$ to 4, as in \cite{Cornernet}). For the predicted interaction vector maps $V$, we use the value of the unsigned interaction vector $v'_{k}=(|v_{x}|_{k},|v_{y}|_{k})$ at the interaction point $p_{k}$ as the ground-truth. Then, L1 loss is employed for all the interaction points,
\begin{equation}
\label{loss_for_vector}
L_{v}=\frac{1}{N}\sum_{k=1}^{N}|V_{p_{k}}-v'_{k}| \,.
\end{equation}
Here $V_{p_{k}}$ denotes the predicted interaction vectors at point $p_{k}$. The overall loss function is summarized as,
\begin{equation}
\label{loss_for_vector}
L_{tot}=L_{p}+\lambda_{v}L_{v} \,,
\end{equation}
where $\lambda_{v}$ is the weight for the vector loss term. Here we simply set $\lambda_{v} = 0.1$ for all our experiments.

\section{Experiments}

\subsection{Datasets and Metrics}
\noindent \textbf{Datasets:}
We conduct comprehensive experiments on two challenging HOI datasets: V-COCO~\cite{gupta2015vcoco} and HICO-DET~\cite{chao2015hico}. The V-COCO dataset contains 2533, 2867, and 4946 images for training, validation and testing, respectively. Typically, the combined  training and validation sets (5400 images in total) are used for model training. Human instances in V-COCO dataset has 26 binary action labels and three action categories (cut, hit, eat) are annotated with two types of targets (i.e., instrument and direct object). Note that three classes (run, stand, walk) are annotated with no interaction object. The
HICO-DET dataset contains 38,118 images for training and 9658 images for testing. In this dataset, each human instance is annotated with 600 classes of different interactions, corresponding to 80 object categories and 117 action verbs. Note that those 117 action verbs include the 'no interaction' class.

\noindent \textbf{Metrics:}
We follow the standard evaluation protocols, as in \cite{gupta2015vcoco, chao2015hico}, to evaluate our proposed method. The results are reported in terms of role mean Average Precision ({mAP$_\text{role} $}). In {mAP$_\text{role} $}, one HOI triplet is regarded as a true positive only when both the human and object detected bounding-boxes have IoUs (intersection-over-union) greater than 0.5 with the respective ground-truth and the associated interaction class is correctly classified.\vspace{-0cm}
\subsection{Implementation Details}
For interaction prediction, we use Hourglass-104 \cite{hourglass} as feature extractor, pre-trained on MS COCO (train2017 set), as in \cite{Zhou&CenterNet}. The head network for the interaction point and interaction vector generation is randomly initialized. During training, we adopt an input resolution of  $512 \times 512$. This yields an output resolution of $128 \times 128$ for the Hourglass backbone. We employ standard data augmentation techniques (random flip, random scaling (between 0.6 to 1.3), cropping, and color jittering) and use Adam optimizer \cite{adam2014} to optimize the loss function during training. During test, we use flip augmentation to obtain final predictions. Following~\cite{Zhou&CenterNet}, we use batch-size of 29 (on 5 GPUs, with master GPU batch-size 4) and learning rate $2.5\cdot 10^{-4}$ for 50 epochs with $10\times$ learning rate drop at 40 epoch. 
For detection branch, we follow previous HOI detection methods \cite{gao2018ican, li2019TIN, wang2019iccv} and utilize Faster-RCNN \cite{fasterrcnn2015} with ResNet-50-FPN \cite{lin2017fpn} pre-trained on COCO \cite{mscoco} train2017 split. 
To obtain bounding-boxes at inference, we set score threshold greater than 0.4 for humans and 0.1 for objects. These score thresholds are relatively lower than the thresholds set in \cite{gao2018ican, wang2019iccv}, since the interaction box generated by our  interaction point and vector can filter out most negative pairs. For interaction generation, Hourglass-104 takes about 77ms. Our interaction grouping has a complexity of $\mathcal{O}(N_\text{h}N_\text{o}N_\text{ip})$, where $N_\text{h}$, $N_\text{o}$, $N_\text{ip}$ is the number of humans, objects and interaction points, respectively. In practice, our grouping is efficient, taking less than 5 ms ($< 6.1\%$ of total time).




 \begin{table}[t!]
\begin{center}
\begin{tabular}{l|c}
\hline
Methods   & \textbf{mAP$_\text{role} $} \\
\hline
VSRL\cite{gupta2015vcoco}* & 31.8\\
InteractNet  \cite{gkioxari2017interactnet} & 40.0 \\ 
BAR \cite{bar2018} & 41.1\\
GPNN \cite{qi2018learning} & 44.0\\
iCAN \cite{gao2018ican} & 45.3\\
HOI w knowledge \cite{Xu2019Knowledge} &45.9\\
DCA \cite{wang2019iccv} & 47.3\\
RPNN \cite{zhou2019iccv} & 47.5\\
TIK \cite{li2019TIN} & 47.8\\
PMFNet \cite{PMFNet} & 52.0 \\
\hline
Ours  & \textbf{51.0}\\
Ours + HICO  & \textbf{52.3}\\
\hline
\end{tabular}
\end{center}\vspace{-0.2cm}
\caption{State-of-the-art comparison (in terms of {mAP$_\text{role} $}) on the V-COCO dataset. * refers to implementation of \cite{gupta2015vcoco} by \cite{gkioxari2017interactnet}. Our approach sets a new state-of-the-art with mAP$_\text{role} $ of 51.0 and achieves an absolute gain of 3.2\% over TIK \cite{li2019TIN}. The results are further improved (mAP$_\text{role} $ of 52.3) when utilizing pre-training on HICO-DET and then fine-tuning on V-COCO dataset.} \vspace{-0.2cm}
\label{tab:v-coco-compare}
 \end{table} 
 
 \begin{table}[t]
\begin{center}
\resizebox{\linewidth}{!}{
\begin{tabular}{l|ccc |ccc}
\hline
 &\multicolumn{3}{c|}{Default}  &\multicolumn{3}{c}{ Known Object} \\
\cline{2-7}
Methods & full & rare & non-rare &full & rare & non-rare\\
\hline
Shen \etal, \cite{shen18}  & 6.46 & 4.24 & 7.12 &- &- &- \\
Chao \etal, \cite{chao2018hoi}  & 7.81 & 5.37 & 8.54 &10.41 &8.94 &10.85\\
InteractNet  \cite{gkioxari2017interactnet} & 9.94 & 7.16 & 10.77 &- &- &-\\
GPNN \cite{qi2018learning} & 13.11 & 9.34 & 14.23 &- &- &-\\
Xu et.al \cite{Xu2019Knowledge} & 14.70 & 13.26 & 15.13 &- &- &-\\
iCAN \cite{gao2018ican} & 14.84 & 10.45 & 16.15 & 16.43 & 12.01 &17.75\\
DCA \cite{wang2019iccv} & 16.24   &11.16   &17.75 &17.73 &12.78 &19.21\\
TIK \cite{li2019TIN} \cite{li2019TIN} & 17.03  &13.42  &18.11  &19.17  &15.51  &20.26\\
Gupta et.al \cite{NoFrillHOI} & 17.18  &12.17  &18.68  &-  &-  &-\\ 
RPNN \cite{zhou2019iccv}  & 17.35  &12.78  &18.71  &-  &-  &-\\
PMFNet \cite{PMFNet} & 17.46   &\textbf{15.65}   &18.00 &20.34 &\textbf{17.47} &21.20 \\
Peyre et.al \cite{Peyre_ICCV19} & 19.40   &14.60   &20.90 &- &- &- \\\hline
Ours & \textbf{19.56}   &12.79 &\textbf{21.58} &\textbf{22.05} &15.77 &\textbf{23.92}\\
\hline
\end{tabular}
}
\end{center}\vspace{-0.2cm}
\caption{State-of-the-art comparison (in terms of {mAP$_\text{role} $}) on the HICO-DET using two different settings: Default and Known Object on all three sets (full, rare, non-rare). Note that Shen \etal \cite{shen18}, InteractNet  \cite{gkioxari2017interactnet} and GPNN \cite{qi2018learning} only report results on the Default settings. For both settings, our approach provides superior performance compared to existing methods. In case of default settings, our approach achieves {mAP$_\text{role} $} of 19.56 on the full set. Further, our approach obtains an absolute gain of 2.9\% over TIK \cite{li2019TIN} on the full set of Known Object setting.  
}\vspace{-0.3cm}\label{tab:hico-det-compare}
\end{table}

\subsection{State-of-the-art Comparison:}
We first compare our proposed approach with state-of-the-art methods in literature. Tab.~\ref{tab:v-coco-compare} shows the comparison on the V-COCO dataset. Among existing approaches, BAR \cite{bar2018}, iCAN \cite{gao2018ican} and  DCA~\cite{wang2019iccv} utilize human and object appearance features in a multi-stream architecture. The DCA method~\cite{wang2019iccv} consisting of a deep contextual attention module that generates contextually-aware appearance features within a multi-stream architecture achieves a {mAP$_\text{role} $} of 47.3. The RPNN approach \cite{zhou2019iccv} based on attention graphs for parsing relations of object and human body-parts obtains a {mAP$_\text{role} $} of 47.3. The work of \cite{li2019TIN}, denoted in Tab.~\ref{tab:v-coco-compare} as TIK, introduces an interactiveness network to perform Non-interaction Suppression and reports a {mAP$_\text{role} $} of 47.8. Our approach achieves superior performance compared to existing methods with a {mAP$_\text{role} $} of 51.0. The results are further improved ({mAP$_\text{role} $} of 52.3) by first pre-training our network on HICO-DET and then fine-tuning the pre-trained HICO-DET model on the V-COCO dataset.

 Tab.~\ref{tab:hico-det-compare} shows the comparison on HICO-DET. As in \cite{chao2015hico}, we report results on three different HOI category sets: full, rare, and non-rare with two different settings of Default and Known Objects. Our approach achieves superior performance compared to the state-of-the-art on both settings. For the Default settings, our approach obtains {mAP$_\text{role} $} of 19.56, 12.79 and 21.58 on the full, rare and non-rare sets, respectively. In case of Known Object setting, our approach achieves an absolute gain of 2.9\% over \cite{li2019TIN} on the full set.

\begin{figure*}[t]
\begin{center}
    \includegraphics[width=\linewidth]{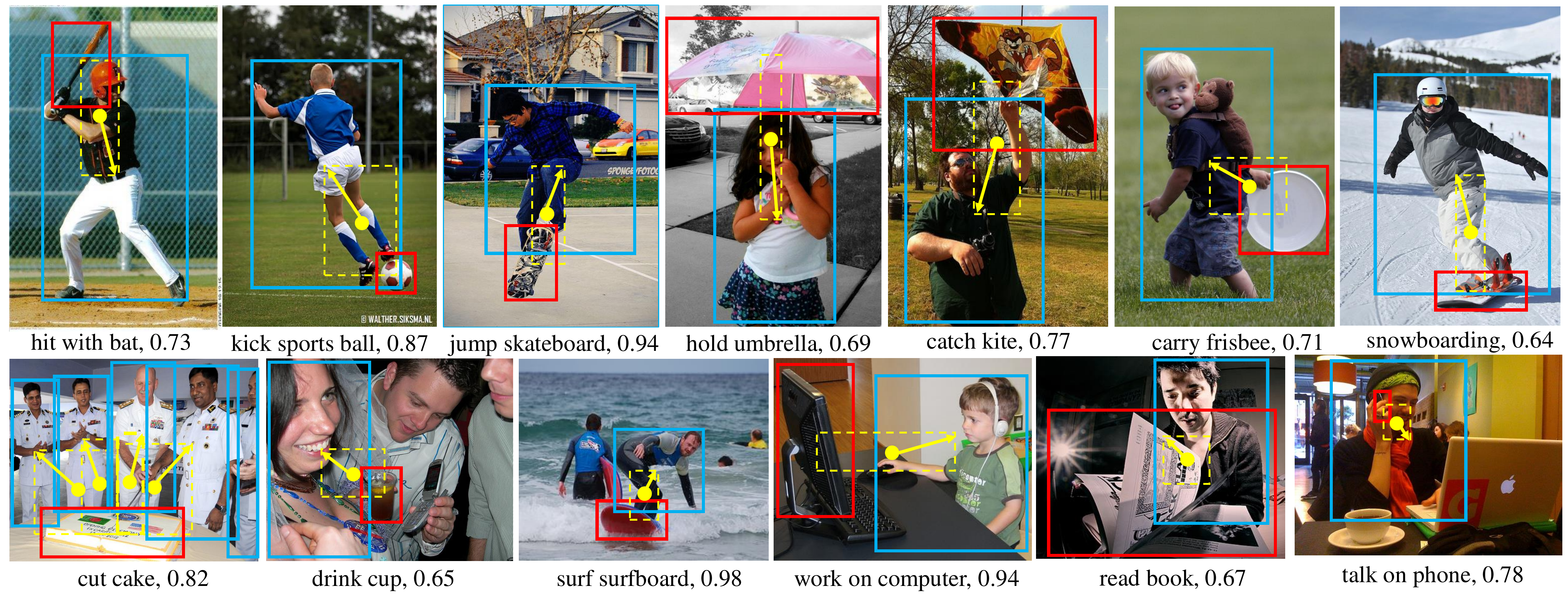}
\end{center} \vspace{-0.5cm}
   \caption{Example detections on V-COCO. Each example involves a human-object interaction, such as \textit{skateboarding} or multiple humans sharing the same interaction and object - \textit{cut cake}. We also show our interaction point, vector and box (in yellow).} \vspace{-0.3cm}
   \label{fig:V-COCO-qualitative}
\end{figure*} 
\subsection{Ablation Study}
We first perform an ablation study on V-COCO using the Hourglass backbone to show the impact of different components in our approach. Tab.~\ref{tab:slice_experiment} shows the impact of interaction points, angle-filter, dist-ratio-filter, interaction boxes, corner distance and center-pool components on the V-COCO dataset. To validate the effectiveness of our interaction grouping scheme, we first compare the proposed interaction grouping with the so-called `angle-filter' and `dist-ratio-filter'. Tab.~\ref{tab:slice_experiment} shows these comparisons on V-COCO. 

\noindent\textbf{Angle Filter: } During training, the interaction point $P$ and its corresponding human center point $H$, object center point $O$ have a fixed geometric structure, \ie, the angle between the vector $\overrightarrow{PH}$ and vector $\overrightarrow{PO}$ is equal to $\pi$. During inference, angle filter aims to reduce those HOI pairs for which the angle between $\overrightarrow{PH}$ and vector $\overrightarrow{PO}$ is lower than a given angle threshold. Tab.~\ref{tab:slice_experiment} shows that this baseline of interaction points with angle filter achieves a \mbox{mAP$_\text{role}$} of 39.6.

\noindent\textbf{Dist-ratio Filter: } Similar to the angle-filter, the ratio between $|\overrightarrow{PH}|$ and $|\overrightarrow{PO}|$ is equal to 1 during training. Therefore the dist-ratio filter can also be employed to filter HOI pairs where the ratio between $\max(|\overrightarrow{PH}|,|\overrightarrow{PO}|)$ and $\min(|\overrightarrow{PH}|,|\overrightarrow{PO}|)$ is greater than the given distance ratio threshold.  Tab.~\ref{tab:slice_experiment} shows this constraint improves over interaction points with angle filter by 1.7\% in terms of {mAP$_\text{role}$}. 

\noindent\textbf{Interaction Grouping:} To explore the effectiveness of our proposed interaction grouping scheme, we divide this scheme into two parts: \noindent\emph{interaction box} and \noindent\emph{corner-dist}, to verify the power of three soft constraints in~\eqref{eq:conditions}. During training, the IoU between the human/object bbox and interaction box is greater than zero. Therefore, to satisfy the first two IoU conditions in \eqref{eq:conditions}, we first integrate the interaction box generated by the interaction vectors to filter out the negative HOI pairs. Tab.~\ref{tab:slice_experiment} shows that it significantly improves the HOI detection performance to 46.2 {mAP$_\text{role} $}. We also found that when only adding interaction box on the interaction points, it further improves the performance by 2\%, from 46.2 to 48.2. Note that, in our approach the four corners of the interaction box are considered as the four corners of the reference box during training. At inference, with the corner distance constraint ($|\text{dist}|<d_{\tau}$) in \eqref{eq:conditions}, some negative pairs are further filtered out resulting in an improved overall performance of 50.5 {mAP$_\text{role}$}.

\begin{table}[t]
\begin{center}
\resizebox{\linewidth}{!}{
\begin{tabular}{lcccccc}
\hline
Add-on  &Baseline  &\multicolumn{5}{c}{}  \\
\hline
\textit{interaction points}   &\checkmark &\checkmark &\checkmark &\checkmark &\checkmark &\checkmark\\
\textit{angle-filter}   &\checkmark &\checkmark &\checkmark &{} &{} &{}\\ 
\textit{dist-ratio-filter}    &{}  &\checkmark &\checkmark &{} &{} &{}\\
\textit{interaction box}  &{} &{}  &\checkmark &\checkmark &\checkmark &\checkmark\\
\textit{corner-dist}   &{} &{}  &{} &{} &\checkmark &\checkmark\\
\textit{center-pool}   &{}  &{} &{}  &{} &{} &\checkmark\\
\hline
mAP$_\text{role} $     &39.6  &41.3 &46.2 &48.2 &50.5 &\textbf{51.0}\\
\hline
\end{tabular}
}
\end{center}\vspace{-0.2cm}
\caption{Impact of integrating our contributions into the baseline on V-COCO. Results are reported in terms of role mean average precision ({mAP$_\text{role}) $}. For fair comparison, we use the same backbone (Hourglass-104) for all the ablation experiments. Our overall architecture achieves a absolute gain of 11.4\% over the baseline.
 }\vspace{-0.3cm}
\label{tab:slice_experiment}
\end{table}

\begin{table}[t!]
\begin{center}
\resizebox{\linewidth}{!}{
\begin{tabular}{l|ccccc|c}
\hline
Score thres &0.01 &0.02  &0.05 &0.08 &0.10 &Dynamic\\
\hline
\textbf{Default:} & &  & & & &\\
Full &19.26 &19.32  &19.08 &18.66 &18.25 &\textbf{19.56}\\
Rare &12.53 &12.00  &10.32 &9.13 &8.12 &\textbf{12.79}\\
Non-rare &21.27 &21.51  &\textbf{21.70} &21.50 &21.27 &21.58\\
\hline
\textbf{Known-Obj:} & &  & & & &\\
Full &21.80 &21.81  &21.57 &21.08 &20.65 &\textbf{22.05}\\
Rare &15.74 &15.06  &13.39 &12.00 &10.79 &\textbf{15.77}\\
Non-rare &23.61 &23.83  &\textbf{24.01} &23.80 &23.60 &23.92\\
\hline
\end{tabular}
}
\end{center}\vspace{-0.2cm}
\caption{Performance comparison (in terms of {mAP$_\text{role} $}) regarding the classification capabilities of our approach for the rare and non-rare classes on the HICO-DET. We show the results with different score thresholds, used during the evaluation. Our proposed dynamic threshold inference achieves a good performance trade-off between the rare and non-rare classes.  }\vspace{-0.3cm}
\label{tab:over_thresh}
\end{table}

\begin{figure*}[t]
\begin{center}
    \includegraphics[width=\linewidth]{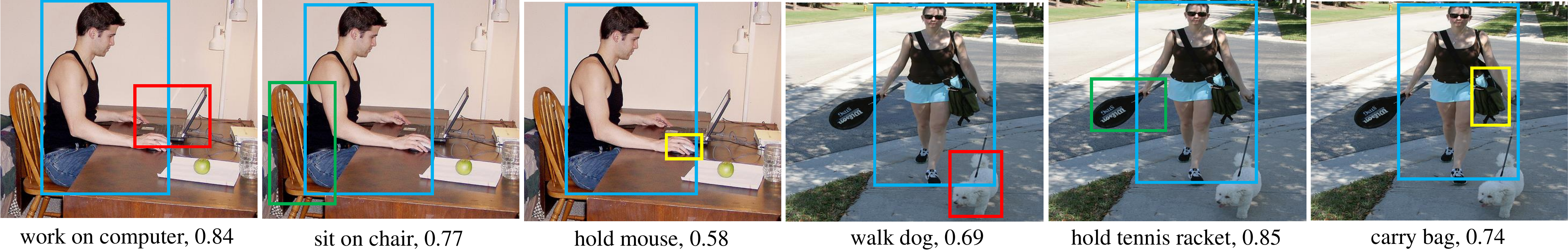}
\end{center} \vspace{-0.4cm}
  \caption{Multiple interaction detection on V-COCO. Our approach detects human instance doing multiple (different) actions and interacting with various objects (represented with different colors). In all cases, the detected agent is represented with the same color.} \vspace{-0.3cm}
  \label{fig:multi-interactions}
\end{figure*} 

\begin{figure*}[t]
\begin{center}
    \includegraphics[width=\linewidth]{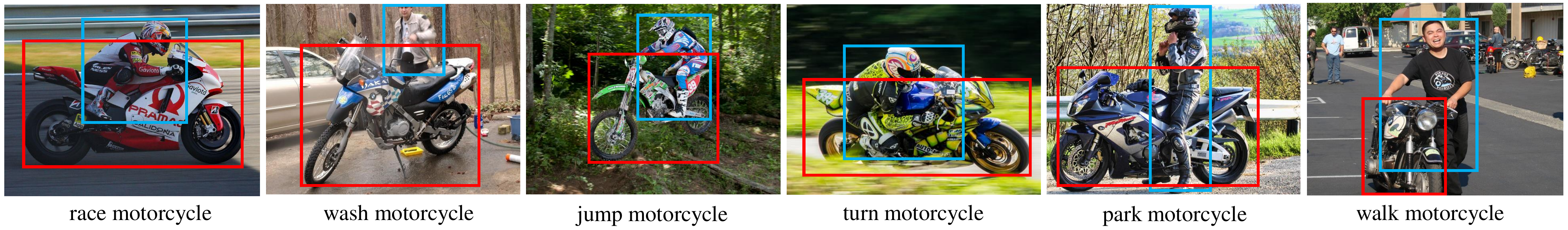}
\end{center} \vspace{-0.3cm}
   \caption{Results on HICO-DET showing one detected triplet. Blue boxes represent a detected human instance, while the red boxes show the detected object of interaction. Our approach detects various fine-grained interactions.}\vspace{-0.2cm} 
   \label{fig:HICO-qualitative}

\end{figure*}



\noindent\textbf{Center-pool:} For improved detection of interaction points, we introduce the center-pool operation, as in in~\cite{Duan&CenterNet}, aiding to obtain more distinct visual patterns between the human and object instances. This operation is achieved by getting out the max summed
response in both horizontal and vertical directions of the interaction point on the feature map. In our method, this is employed before the interaction point and vector branch. This operation results in a slight improvement in performance (0.5 points), as shown in Tab.~\ref{tab:slice_experiment}.

We also conduct experiments to evaluate the impact of interaction score thresold on rare and non-rare action classes on HICO-DET. We select different interaction thresholds in the range $[0.01,0.1]$, used in the test evaluation of interaction recognition performance. The results are presented in Tab.~\ref{tab:over_thresh}. These results suggest that the suitable score thresholds are different for the rare and non-rare interaction classes. This is likely due to the fact that the rare classes include less training samples, which results in relatively lower prediction scores for those classes. In contrast, the prediction scores for the non-rare classes tend to be relatively higher. Therefore, it becomes a trade-off problem for the point-based HOI methods to obtain a good performance. We further develop a dynamic threshold inference, which sets different score thresholds for different interaction classes based on their training samples. As show in Tab.~\ref{tab:over_thresh}, our dynamic threshold inference leads to a good performance trade-off between rare and non-rare classes.

\subsection{Qualitative Visualization Results}
\begin{figure}[t]
\centering\vspace{-3mm}%
   \includegraphics[width=0.48\textwidth]{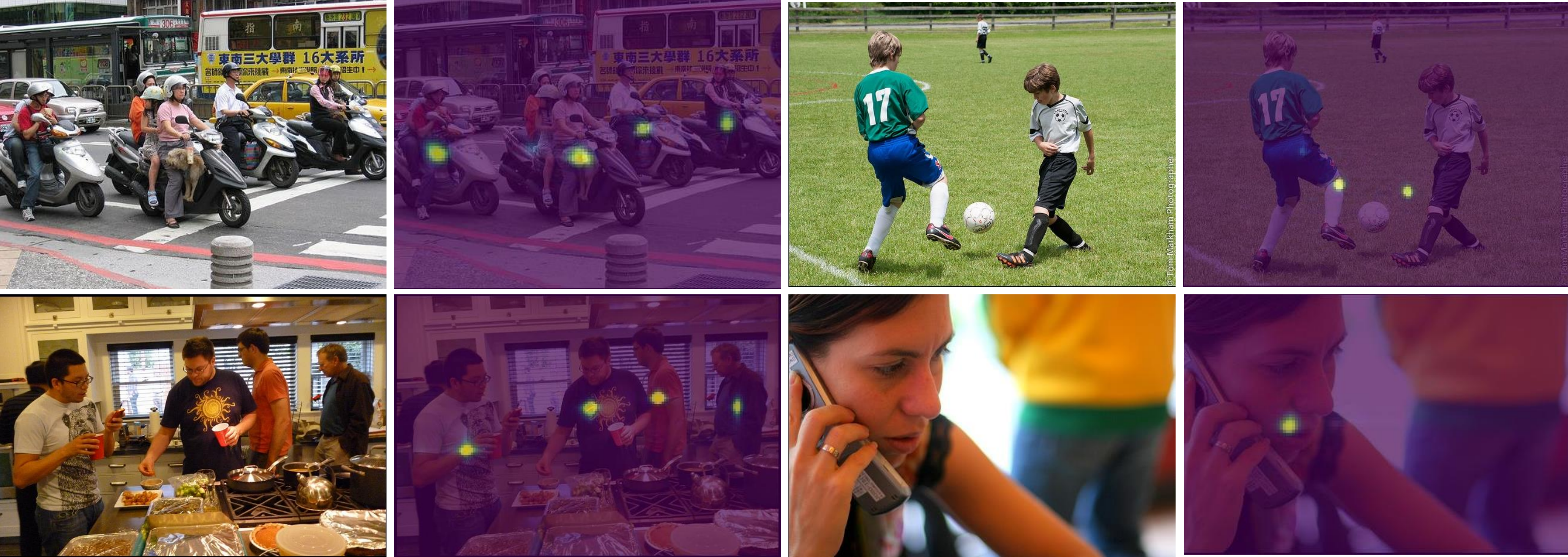}
 \vspace{-0.3cm}
   \caption{Visualization of interaction point heat-maps. Our method is able to cope with challenging scenarios, such as multiple HOI pairs and multiple humans sharing the same object.}
 \vspace{-0.3cm}
\label{feature_vis}
\end{figure}

Fig.~\ref{fig:V-COCO-qualitative} shows examples of both single human-object interactions, such as \textit{hold a umbrella} and \textit{work on computer}, and multiple humans sharing same interaction and object (\textit{cut cake}) along with corresponding interaction scores on V-COCO. The interaction boxes (yellow dash line) generated by the interaction vectors (yellow solid line) are also drawn. These interaction boxes are paired with the positive human and object bounding-boxes using interaction grouping. Fig.~\ref{fig:multi-interactions} shows examples of a human performing multiple interactions. Different interaction objects are annotated with bounding-boxes of different colors. 
Fig.~\ref{fig:HICO-qualitative} shows fine-grained human-object interaction results on HICO-DET. The heatmap visualization for the interaction point map is shown in Fig.~\ref{feature_vis}. Similar to previous works, we observe long-tailed classes to be particularly challenging for HOI detection. Further,  a minor limitation of our approach is that multiple HOI pairs cannot share the same interaction point. However, such cases are rare in practice.


\section{Conclusion}
We propose a point-based framework for  HOI detection. Our approach regards the HOI detection as a keypoint detection and grouping problem. The interaction point and its corresponding interaction vector are first generated by the keypoint detection network. Then, we directly 
pair those interaction points with the human and object bounding boxes from object detection branch using the proposed interaction grouping scheme. Experiments are performed on two HOI detection benchmarks. Our points-based approach outperforms state-of-the-art methods on both datasets.





{\small
\bibliographystyle{ieee_fullname}
\bibliography{egbib}
}

\end{document}